\documentclass{article}
\usepackage{amssymb}
\usepackage{amsmath}
\usepackage{algorithm}
\usepackage{algpseudocode}
\usepackage{booktabs}
\usepackage{subcaption}
\usepackage{float}
\usepackage{graphicx}
\usepackage[preprint]{corl_2025} 

\title{Selective Progress-Aware Querying for Human-in-the-Loop Reinforcement Learning}

\author{
  Anujith Muraleedharan, Anamika J H\\
}

\begin{document}
\maketitle


\begin{abstract}
    Human feedback can greatly accelerate robot learning, but in real-world settings, such feedback is costly and limited. Existing human-in-the-loop reinforcement learning (HiL-RL) methods often assume abundant feedback, limiting their practicality for physical robot deployment. In this work, we introduce SPARQ, a progress-aware query policy that requests feedback only when learning stagnates or worsens, thereby reducing unnecessary oracle calls. We evaluate SPARQ on a simulated UR5 cube-picking task in PyBullet, comparing against three baselines: no feedback, random querying, and always querying. Our experiments show that SPARQ achieves near-perfect task success, matching the performance of always querying while consuming about half the feedback budget. It also provides more stable and efficient learning than random querying, and significantly improves over training without feedback. These findings suggest that selective, progress-based query strategies can make HiL-RL more efficient and scalable for robots operating under realistic human effort constraints.
\end{abstract}



\section{Introduction}
	
 Robots deployed in the real world must adapt to diverse and dynamic environments while operating under safety and efficiency constraints. HiL-RL has emerged as a powerful paradigm to align robot behavior with human intent by incorporating interactive signals such as evaluative feedback \citep{knox2008tamer}, corrective actions\citep{warnell2018deep,perez2018dcoach}, and preference comparisons \citep{christiano2017deep}. These approaches accelerate learning in tasks ranging from manipulation to navigation \citep{hoque2022thriftydagger,honeycutt2020soliciting}. However, their practicality is limited by the cost of human supervision: attention is a scarce resource, constrained by fatigue, multitasking demands, and operational limitations \citep{honeycutt2020soliciting,heo2020costeffective}. Continuous querying for feedback overwhelms supervisors, while too few queries slow adaptation and degrade performance.
 
To address this tension, interactive strategies have been proposed where robots query humans only when needed. “Human-gated” approaches allow supervisors to intervene when they see fit \citep{perez2018dcoach,celemin2019coach}, but require continuous monitoring and cannot scale to multi-robot settings. “Robot-gated” methods shift the responsibility to the agent, enabling it to request feedback when encountering novel or risky states \citep{wirth2017survey,kaufmann2023survey}. While these strategies reduce unnecessary interactions, they typically lack explicit mechanisms for managing strict feedback budgets and often struggle in continuous-control tasks where queries must be carefully timed to avoid disrupting smooth execution.

\begin{figure}[t]
  \centering
  \includegraphics[width=\linewidth]{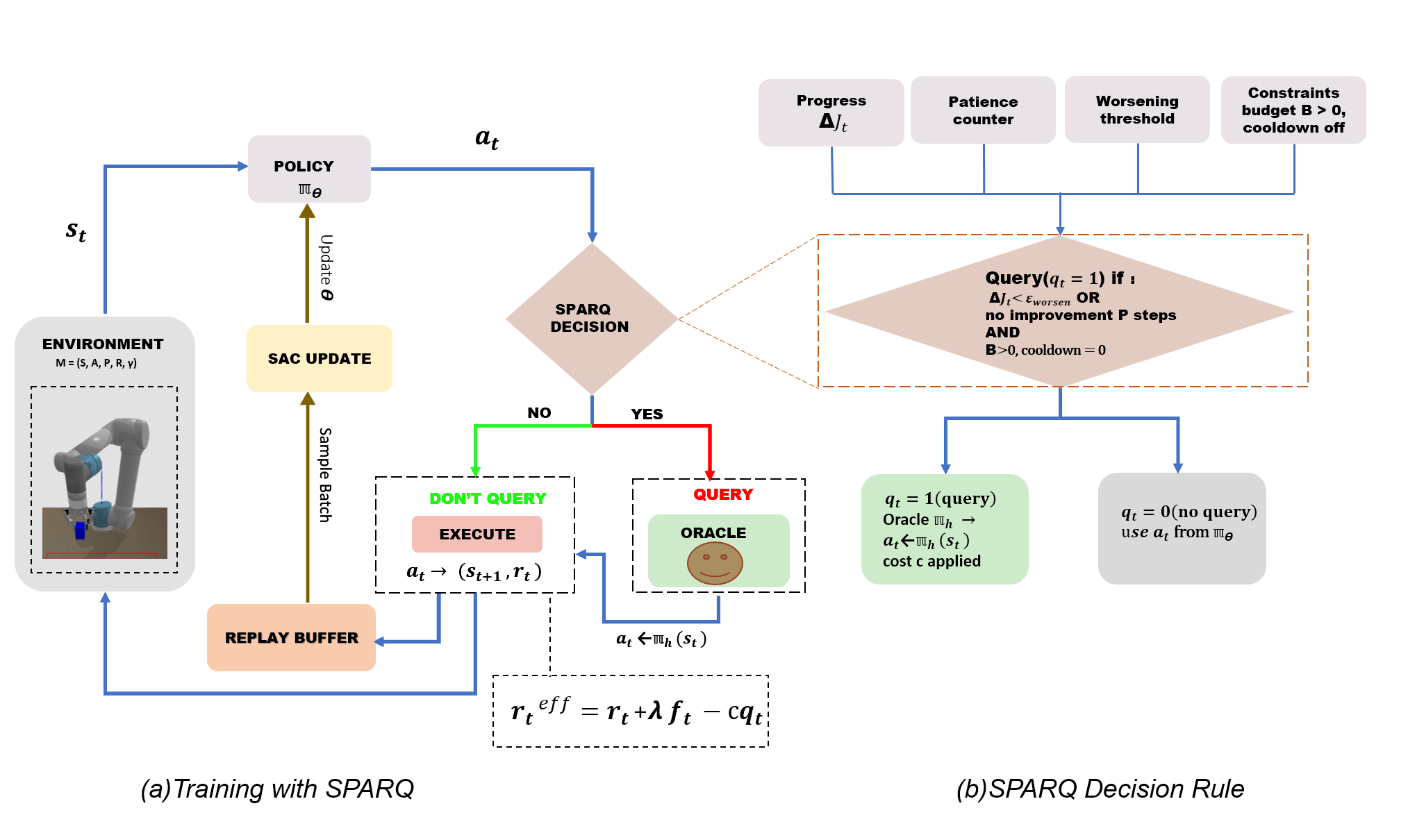}
  \caption{
  (a) \textbf{Training with SPARQ:} the policy $\pi_\theta$ samples $a_t$, the SPARQ gate decides whether to query the human oracle $\pi_h$; the executed transition $(s_t,a_t,s_{t+1},r_t)$ is stored with the effective reward $r_t^{\mathrm{eff}}=r_t+\lambda f_t-c q_t$ in replay $D$, and SAC updates $\theta$.
  (b) \textbf{SPARQ decision rule (zoom-in):} query when $(\Delta J_t<-\epsilon_{\text{worsen}})$ or (no improvement for $P$ steps), and only if $B>0$ and cooldown$=0$; otherwise use $a_t$ from $\pi_\theta$.
  }
  \label{fig:sparq-overview}
\end{figure}

Inspired by how humans naturally allocate their effort rationally, we propose SPARQ (Selective Progress-Aware Querying for Human-in-the-Loop Reinforcement Learning), a budget-aware HiL-RL method. Rather than relying on continuous or uncertainty-driven feedback, SPARQ monitors task progress and selectively requests help only when learning stagnates or worsens. By explicitly modeling human attention as a limited budget and enforcing cooldowns between queries, SPARQ balances learning efficiency with supervision cost. An overview of our approach is illustrated in Fig.~\ref{fig:sparq-overview}. Panel (a) shows the SPARQ-augmented training pipeline, where the agent selectively queries a human oracle only when needed. Panel (b) zooms into the SPARQ decision rule, which determines when queries are triggered based on progress, patience, and budget constraints.

Our contributions are as follows:
\begin{itemize}
   \item We propose SPARQ, a resource-rational HiL-RL method that models human attention as a budgeted resource and allocates queries selectively.
   \item We introduce a progress-aware query rule that triggers feedback requests based on learning stagnation or deterioration, with patience and cooldown to prevent redundancy.
   \item We empirically evaluate SPARQ on a simulated UR5 cube-picking task, showing that it matches the success rate of full-feedback baselines while using about half the feedback budget, and performs comparably to Random on this task while providing explicit budget control and smoother temporal allocation of queries.
\end{itemize}


\section{Related Work}
\label{sec:Related Work}
    
Human-in-the-loop reinforcement learning (HiL-RL) has been studied extensively through frameworks that leverage explicit human feedback to accelerate agent learning. Early approaches such as TAMER\citep{knox2008tamer} and its extension Deep TAMER\citep{warnell2018deep} enabled agents to learn policies directly from human reward signals, but they required frequent intervention, making them impractical in resource-constrained scenarios. PEBBLE \citep{lee2021pebble} improved upon this by using preference-based feedback to reduce cognitive burden, yet it assumes relatively abundant annotations. Similarly, COACH \citep{celemin2019coach} modeled human feedback as policy-dependent, but lacked scalability to complex, high-dimensional tasks. These methods underscore the benefits of human guidance but fail to explicitly account for query efficiency, often leading to redundant human effort.

On the other hand, imitation learning has explored methods to minimize human supervision while ensuring robust policy transfer. DAgger \citep{ross2011dagger} introduced iterative aggregation of expert demonstrations to mitigate compounding errors, but at the cost of requiring frequent expert corrections. HG-DAgger \citep{kelly2019hgdagger} extended this paradigm by letting a human supervisor decide when to provide demonstrations, reducing the labeling load compared to vanilla DAgger. However, this “human-gated” mechanism still relies on continuous monitoring, limiting scalability in practice. To address this, ThriftyDAgger \citep{hoque2022thriftydagger} proposed querying experts only at uncertain states, but its heuristic query strategy lacked robustness in continuous-control settings. More recently, Diff-DAgger \citep{lee2024diffdagger} incorporated diffusion-based uncertainty modeling to refine query selection, yet its reliance on dense expert demonstrations limits its practical applicability in settings with costly supervision.

Our work differs in two key respects. First, unlike TAMER- and PEBBLE-style methods that assume abundant human feedback, SPARQ explicitly models human attention as a limited resource and enforces a feedback budget. Second, unlike DAgger-style approaches that depend on dense expert demonstrations and uncertainty heuristics, SPARQ introduces a progress-aware query criterion: the agent requests feedback only when learning stagnates or deteriorates, combined with patience and cooldown to avoid redundancy. This budget-aware design enables SPARQ to match the performance of full-feedback baselines while cutting feedback usage nearly in half, offering a scalable pathway for HiL-RL in real-world robotics.
	

\section{Methods}
\label{sec:methods}

Our objective is to study query policies that enable efficient human-in-the-loop reinforcement learning (HiL-RL) under limited and costly feedback. 
We first formalize the problem setting as a constrained reinforcement learning task, and then describe four query strategies: three baselines and our proposed approach SPARQ (Selective Progress-Aware Querying for Human-in-the-Loop Reinforcement Learning).

\subsection{Problem Formulation}

We model the robot learning task as a discrete-time Markov Decision Process (MDP), $M = (\mathcal{S}, \mathcal{A}, P, R, \gamma)$ with continuous state space $s \in \mathcal{S}$, continuous action space $a \in \mathcal{A}$, unknown transition dynamics $P$, reward function $R$, and discount factor $\gamma \in [0,1]$.  
The agent’s objective is to learn a policy $\pi_\theta: \mathcal{S} \to \mathcal{A}$ that maximizes expected discounted return. Unlike standard RL, the reward function $R$ is partially specified and can be augmented by human feedback.  
We assume access to a human oracle $\pi_h$ that can provide corrective signals when queried \citep{ross2011dagger,kelly2019hgdagger}. Each query incurs a cost $c > 0$, reflecting the limited time and attention of the human.  

We define the agent’s effective reward at timestep $t$ as:
\begin{equation}
r_t^{\text{eff}} = r_t + \lambda \cdot f_t - c \cdot q_t,
\label{eq:effective-reward}
\end{equation}
where $r_t$ is the environment reward, $f_t$ is the corrective feedback from the oracle $\pi_h$ when queried, $\lambda$ is a scaling factor, and $q_t \in \{0,1\}$ indicates whether a query is made at time $t$.  
This formulation follows potential-based reward shaping principles, which preserve policy invariance \citep{ng1999policy}.  

The total expected return under policy $\pi$ and query policy $\pi_q$ is:
\begin{equation}
J(\pi, \pi_q) = \mathbb{E} \Bigg[ \sum_{t=0}^T \gamma^t \big( r_t + \lambda f_t - c q_t \big) \Bigg].
\label{eq:expected-return}
\end{equation}

We further define $\Delta J_t$ as a proxy for learning progress, e.g., the change in average episodic return or a task-specific metric such as distance-to-goal.  

The goal is to jointly optimize the robot policy and query policy such that the agent learns efficiently while minimizing human effort:
\begin{equation}
(\pi^\ast, \pi_q^\ast) = \arg\max_{\pi, \pi_q} J(\pi, \pi_q),
\label{eq:joint-opt}
\end{equation}
subject to a feedback budget
\begin{equation}
\sum_{t=0}^T q_t \leq B,
\label{eq:budget}
\end{equation}
where $B$ represents the maximum number of queries available.  

The query policy $\pi_q: \mathcal{S} \to \{0,1\}$ thus decides when to request feedback, ideally only when learning progress stagnates or worsens.

\subsection{Baselines}
All query strategies are trained using Soft Actor-Critic (SAC) \citep{haarnoja2018sac}, chosen for its stability and sample efficiency in continuous control tasks. We compare three baselines against our proposed method.

\textbf{No Oracle.} The no-oracle baseline represents standard reinforcement learning without human intervention. The query policy is identically zero, $\pi_q(s) = 0 \;\; \forall s \in \mathcal{S}$, and the effective reward reduces to $r_t^{\text{eff}} = r_t$. This baseline provides a lower bound on sample efficiency.  

\textbf{Random Querying.} In this baseline, the agent queries the oracle with fixed probability $p$ at each timestep: $\pi_q(s) \sim \text{Bernoulli}(p)$. This strategy ignores task structure and may waste queries on uninformative states. It serves as a mid-point between no supervision and full supervision.  

\textbf{Always Querying.} Here, the agent queries the oracle at every timestep: $\pi_q(s) = 1 \;\; \forall s \in \mathcal{S}$. This baseline achieves fast learning but immediately exhausts the query budget and imposes maximal human cost. It functions as an upper bound for performance under unlimited supervision.  

\subsection{SPARQ: Selective Progress-Aware Querying for Human-in-the-Loop Reinforcement Learning}
\begin{algorithm}[!b]
\caption{SPARQ: Selective Progress-Aware Querying for Human-in-the-Loop Reinforcement Learning}
\label{alg:sparq}
\begin{algorithmic}[1]
\Require Initial policy parameters $\theta$, query budget $B$, thresholds $\epsilon_{\text{worsen}}, P$, cooldown $C$
\Ensure Learned policy $\pi_\theta$
\State Initialize cooldown counter $c \gets 0$
\For{each episode}
    \State Reset environment, obtain $s_0$
    \For{each timestep $t = 0, \dots, T$}
        \State Sample $a_t \sim \pi_\theta(\cdot \mid s_t)$
        \State Estimate progress $\Delta J_t$
        \If{$c = 0$ \textbf{and} $B > 0$ \textbf{and} ($\Delta J_t < -\epsilon_{\text{worsen}}$ \textbf{or} no improvement for $P$ steps)}
            \State Query oracle: $a_t \leftarrow \pi_h(s_t)$
            \State Set $q_t \gets 1$, update $B \gets B - 1$
            \State Reset cooldown counter: $c \gets C$
        \Else
            \State Execute $a_t$, set $q_t \gets 0$
            \If{$c > 0$} \State Decrement cooldown counter: $c \gets c - 1$ \EndIf
        \EndIf
        \State Observe reward $r_t$, construct $r_t^{\text{eff}} = r_t + \lambda f_t - c q_t$ \Comment{Eq.~\ref{eq:effective-reward}}
        \State Update $\pi_\theta$ with SAC using $r_t^{\text{eff}}$
        \State Transition to $s_{t+1}$
    \EndFor
\EndFor
\end{algorithmic}
\end{algorithm}
We propose a resource-rational query policy that adaptively balances performance improvement against feedback cost. The key idea is to trigger queries selectively, only when learning progress stagnates or worsens.
  
\textbf{Progress-Aware Criterion.} We also track learning progress using $\Delta J_t$ based on episodic return or distance-to-goal. Queries are triggered when performance worsens beyond $\epsilon_{\text{worsen}}$, or when no improvement is observed for $P$ steps: $\Delta J_t < -\epsilon_{\text{worsen}} \;\lor\; \text{no improvement for $P$ steps}$.  

\textbf{Budget and Cooldown.} To respect the constraint $\sum_{t=0}^T q_t \leq B$, we maintain a finite query budget and impose a cooldown period after each query to avoid redundancy.
Algorithm~\ref{alg:sparq} summarizes the overall procedure for our proposed method SPARQ.

\textbf{Practical tuning.} We tune three scalars once per domain: patience \(P\) (steps without improvement), worsening threshold \(\epsilon_{\text{worsen}}\) (minimum negative \(\Delta J_t\) to trigger), and cooldown \(C\) (steps to defer re-queries). 
A simple guideline is: choose \(P\) near \(1\text{--}2\times\) the median episode length; set \(\epsilon_{\text{worsen}}\) to the \(5\text{--}10\)th percentile of negative \(|\Delta J_t|\) magnitudes observed early in training; and pick \(C\) so the average query rate tracks the target feedback budget (e.g., adjust \(C\) until \(\frac{1}{T}\sum_{t=1}^T q_t \approx b/100\) for a desired \(b\%\)). These rules-of-thumb yielded stable behavior without per-seed retuning in our setting.


\section{Experimental Results}
\label{sec:result}

We evaluate SPARQ against three baseline query policies---No Oracle, Random Querying, and Always Querying, on a simulated UR5 cube-picking task in PyBullet. The evaluation addresses the following research questions:

\begin{itemize}
    \item \textbf{RQ1:} Can progress-aware querying improve query efficiency under a limited budget?
    \item \textbf{RQ2:} Does SPARQ achieve comparable or better task success rates than baselines with fewer queries?
    \item \textbf{RQ3:} How do different query policies trade off task performance and human effort cost?
\end{itemize}

\subsection{Task Setup}

The evaluation task requires the UR5 arm to reach and grasp a cube placed at random positions on a tabletop workspace.  
The action space consists of continuous $(x,y)$ end-effector target positions, while the observation space is the cube’s $(x,y)$ location.  
A sparse success reward of $+1$ is given when the cube is successfully grasped, augmented with potential-based shaping based on distance-to-goal. All query policies are trained using Soft Actor-Critic (SAC) for $50$k timesteps. Hyperparameters such as the feedback budget $B$, query cost $c$, and patience $P$ are fixed across all methods.

\subsection{Query Efficiency (RQ1)}

An effective query policy should minimize unnecessary oracle requests while still providing sufficient supervision to enable learning.  
We measure query efficiency as the average number of oracle requests per episode relative to the nominal budget.  

As shown in Table~\ref{tab:combined} and Fig.~\ref{fig:query-usage}, SPARQ consumes $\sim 13\%$ of the budget, compared to ~27\% for Always Querying, effectively halving supervision demand. 
This demonstrates that progress-aware triggering avoids redundant queries that random or unconditional strategies incur. We allow occasional post-convergence queries due to replay shaping and drift checks; in deployment, a stop-when-converged guard (zero queries once success \(\ge\) a threshold over a patience window) would eliminate these residual queries.

\begin{table}[H]
\centering
\small
\setlength{\tabcolsep}{6pt}
\begin{tabular}{lccccc}
\toprule
Method & Success Rate & Budget \% & Cost-Adj. Return & Queries / Success & Final Dist \\
\midrule
always     & $1.000 \pm 0.000$ & 26.6\% & 198.4 & 1.33 & $0.001 \pm 0.000$ \\
no\_oracle & $0.610 \pm 0.490$ & 0.0\%  & 91.1  & 0.00 & $0.002 \pm 0.000$ \\
random     & $1.000 \pm 0.000$ & 12.6\% & 196.6 & 0.63 & $0.003 \pm 0.001$ \\
sparq      & $1.000 \pm 0.000$ & 13.2\% & 192.4 & 0.66 & $0.002 \pm 0.001$ \\
\bottomrule
\end{tabular}
\vspace{3pt}
\caption{Overall comparison at 50k steps (mean $\pm$ std unless noted). 
Cost-Adj.\ Return assumes query cost $c=0.05$. 
Final Dist is reported as median $\pm$ MAD end-effector$\to$cube distance at grasp termination.}
\label{tab:combined}
\end{table}
\subsection{Task Performance (RQ2)}

\begin{figure}[H]
  \centering
  \begin{subfigure}{0.49\linewidth}
    \includegraphics[width=\linewidth]{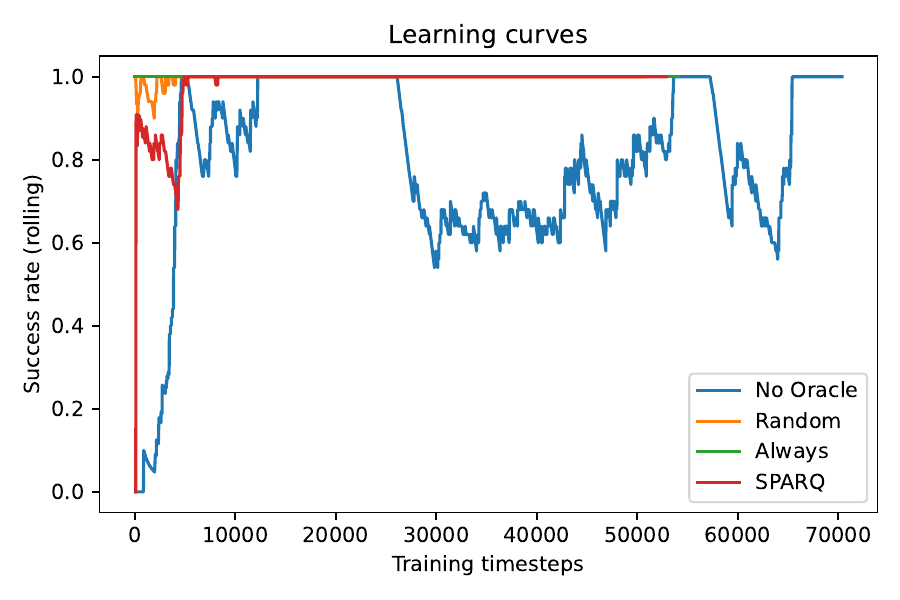}
    \caption{Learning curves (success vs. timesteps).}
    \label{fig:learning-curves}
  \end{subfigure}
  \hfill
  \begin{subfigure}{0.49\linewidth}
    \includegraphics[width=\linewidth]{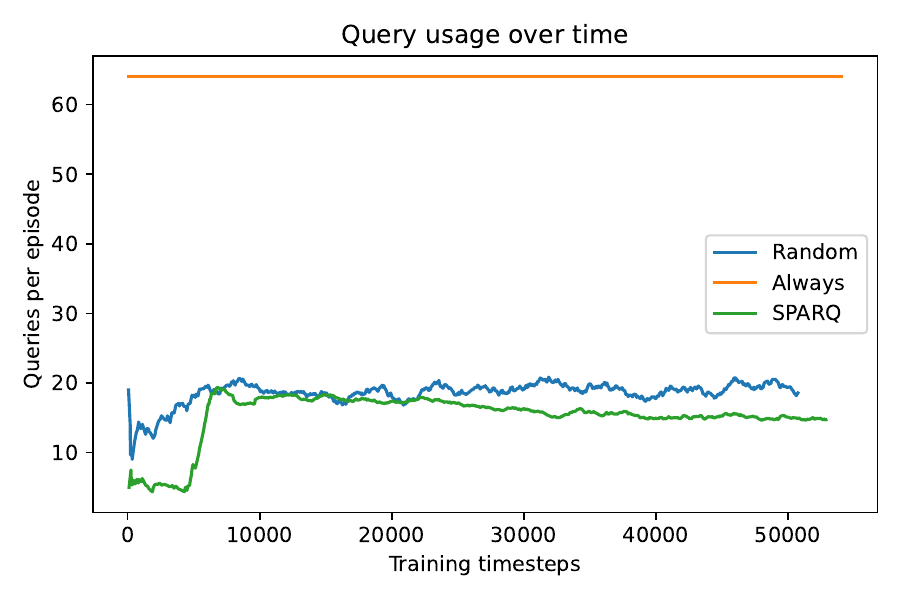}
    \caption{Queries per episode vs. timesteps.}
    \label{fig:query-usage}
  \end{subfigure}
  \vspace{-4pt}
  \caption{Training dynamics on the UR5 cube-picking task. (a) SPARQ converges to $100\%$ success with far fewer queries than Always. (b) SPARQ’s query rate remains low and stable over training.}
  \label{fig:dynamics}
\end{figure}

We next evaluate task success and learning dynamics.  
Figure~\ref{fig:learning-curves} reports success rates and episodic returns across training.  
SPARQ attains a $100\%$ final success rate, on par with Always Querying and substantially higher than No Oracle (61\%).  
Unlike Random Querying, SPARQ achieves faster convergence and more stable training, indicating that its progress-based criterion provides timely and informative feedback.  

\subsection{Trade-off Analysis (RQ3)}

Finally, we examine the trade-off between task performance and human effort cost.  
Table~\ref{tab:combined} summarizes this balance by reporting success rate, budget usage, cost-adjusted return, query efficiency, and final grasp distance.  
SPARQ achieves near-optimal success rates while using only about half as many queries as Always Querying.  
Its cost-adjusted return is nearly identical to the unconstrained baseline, despite consuming substantially fewer human interactions.  
Moreover, SPARQ’s final grasp distance is comparable to or better than other baselines, indicating that reduced supervision does not compromise execution precision.
\subsection{Why does Random perform so well here?}
\label{subsec:why-random}
This benchmark is intentionally simple: a low-dimensional state \((x,y)\) for the cube, a shaped progress signal (distance-to-goal), and stable SAC training. Under these conditions, many query placements are similarly helpful, so a fixed-rate Random policy can sit close to the Pareto front. SPARQ’s advantage shows up in budget regularization and temporal smoothness (Fig.~\ref{fig:dynamics}b): its query rate remains low and steady, avoiding bursts that can fatigue human operators, while achieving the same final success. We view this as evidence that progress-aware gating is a principled default when scaling to higher-dimensional observations and noisier progress proxies.

\section{Conclusion}
\label{sec:conclusion}

	We introduced SPARQ, a resource-rational query framework for budget-aware human-in-the-loop reinforcement learning (HiL-RL). SPARQ monitors learning progress and selectively queries for feedback only when performance stagnates or worsens, ensuring that limited supervision is allocated where it is most impactful. In experiments on a simulated UR5 cube-picking task, SPARQ achieved 100\% final success, matching the unconstrained Always Querying baseline while consuming about half the supervision budget. It also maintained comparable cost-adjusted returns and demonstrated precise grasp accuracy, underscoring that targeted feedback can deliver both efficiency and performance. While these results highlight SPARQ’s potential for scalable HiL-RL, our evaluation is limited to a single simulated domain. A multi-seed robustness study and deployment on real-world robots remain important future directions. Extending SPARQ to incorporate richer forms of human input (e.g., preferences, demonstrations) and to coordinate feedback across multiple robots are natural next steps toward practical, resource-rational HiL-RL.


\clearpage
\acknowledgments{We thank the anonymous reviewers for their valuable feedback and the CoRL workshop organizers for providing a venue to share this work.
}


\bibliographystyle{corlabbrvnat}
\bibliography{example}  

\end{document}